\documentclass[letterpaper, 10 pt, conference]{ieeeconf}  

\IEEEoverridecommandlockouts                              
\usepackage{amsmath} 
\usepackage{graphicx}
\usepackage{amssymb}
\usepackage[latin9]{inputenc}
\usepackage{verbatim}
\usepackage[english]{babel}
\usepackage{subcaption}
\usepackage[table,xcdraw]{xcolor}
\usepackage[obeyFinal,colorinlistoftodos,textsize=small]{todonotes}   
\usepackage[table,xcdraw]{xcolor}

\title{\LARGE \bf Forward and Inverse Kinematics of a Single Section Inextensible Continuum Arm}

\author{Ali A. Nazari, Diego Castro, and Isuru S. Godage
\thanks{The authors are with the School of Computing, DePaul University, Chicago, IL 60604
        {\tt\small anazari@mail.depaul.edu}}%
\thanks{This work was partially supported by National Science Foundation (NSF) grant no. 1718755}
}

\begin{document}

\maketitle
\thispagestyle{empty}
\pagestyle{empty}

\begin{abstract}
Continuum arms, such as trunk and tentacle robots, lie between the two extremities of rigid and soft robots and promise to capture the \textit{best} of both worlds in terms of manipulability, dexterity, and compliance. This paper 
proposes a new 
kinematic model for a novel constant-length continuum robot that incorporates both soft and rigid elements. 
In contrast to traditional pneumatically actuated, variable-length continuum arms, the proposed design utilizes a hyper-redundant rigid chain to provide extra structural strength.
The proposed model introduces a reduced-order mapping to 
account for mechanical constraints arising from the rigid-linked chain to derive a closed-form curve parametric model. The  model is
numerically evaluated and the results show that the derived model is reliable.
\end{abstract}

\section{Introduction}\label{sec:intro}
Continuum robotics has been a highly active area of research in the past few years \cite{walker2013sota}. Continuum arms are typically powered by bio-inspired variable-length actuators, called pneumatic muscle actuators (PMA), which can continuously change their shape using a few actuated degrees of freedom (DoF)  \cite{rus2015design}. PMAs facilitate the formation of complex \textit{organic} shapes in contrast to the fixed \textit{geometric} shapes of rigid-body robotic manipulators \cite{noritsugu1997rehabold}. Continuum arms, inspired by appendages \cite{robinson1999continuum} and elephant trunks \cite{queisser2014active}, have been prototyped over the years. They have been used to demonstrate the potential for adaptive whole arm grasping \cite{braganza2006whole}. They also exhibit improved performance in the areas of obstacle avoidance and compliant operation \cite{li2016progressive}, navigation in highly unstructured, narrow and obstructive environments \cite{godage2012path}, and human-friendly interaction \cite{xiao2010real}. Application areas of continuum arms include, but are not limited to, various medical procedures \cite{burgner2015continuum}.
 
To date, macro-scale pneumatically actuated continuum arms have relied on PMAs for both actuation power and structural integrity \cite{godage2015modal}. This approach has resulted in compliant operation but produced an inferior performance in payload handling \cite{mcmahan2006field}. 
\begin{figure}
\begin{centering}
\includegraphics[width=1\columnwidth]{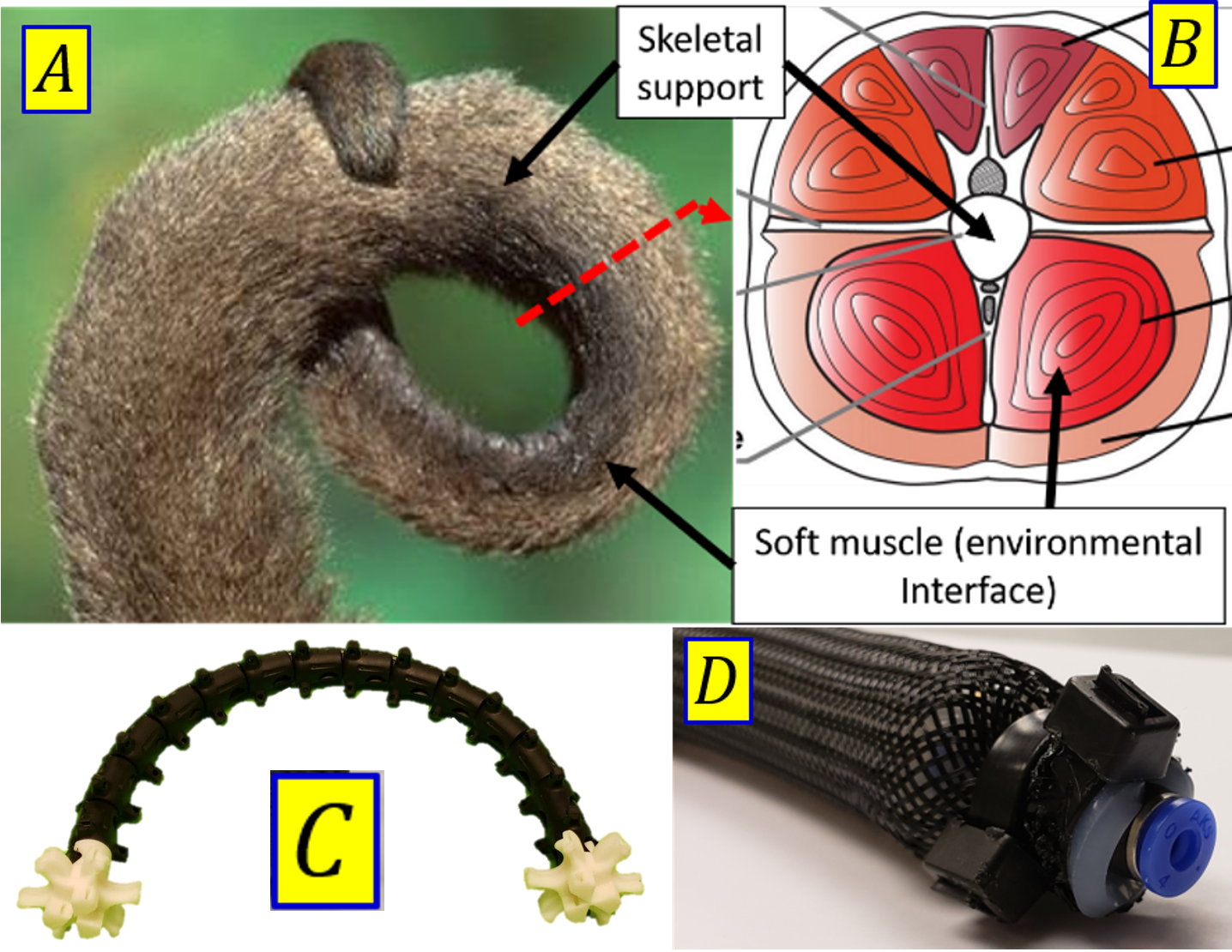}
\par\end{centering}
\caption{(A) Tail of a monkey, (B) Transverse cut of a monkey tail; arm components, (C) 3D-printed backbone, (D) Pneumatic muscle actuator.}
\label{fig:fig1}
\end{figure}
%
In addition, they lack the capability to control either the arm's tip position independent of the arm's length \cite{giannaccini2018novel} or the arm's stiffness independent of the arm's tip position \cite{allaix2018stiffnessAp}.

The implications of the current limitations are twofold; limited actuation scope (either biased toward human-friendly operations) or object manipulation. In the case of the former, for applications like rehabilitation, continuum arms are still expected to be able to handle significant loading. In the case of the human-friendly operation, the current implementation of continuum arms exhibits unpredictable deformation (for example, torsional deformation) or buckling \cite{mustaza2018stiffness}. In order to meet both application requirements, a fluid-based continuum robotic structure is proposed that employs a rigid-link, highly articulated inextensible (that is, constant-length) backbone. The primary motivations of the proposition are 1) to combine stiff and soft elements in the design, enabling the arm to be compliant and human-friendly at the same time structurally solid for high payload manipulation tasks; 2) to increase the stiffness range. As the inextensible backbone introduces an additional constraint, fluid-based muscles can leverage the high power and passive deformation to achieve higher stiffness range \cite{santiago2016soft,mcmahan2006field}; and 3) to make the arm's tip position independent of the arm's length, because of the additional constraint. As such, there is a need to model these complex shapes that are inherent in continuum arms. This necessity has been faced from multiple fronts, each with its own set of benefits and challenges \cite{sadati2017mechanics}.

Early models 
denoted continuum arm as a serially jointed
%
rigid body \cite{robinson1999continuum,giri2011continuum}. Such lumped parameter models for continuum arms mark 
the natural extension from the rigid robotics;
however, they require a large number of jointed links to successfully represent smooth bending. 
 %
Constant-curvature models eliminated the limitations regarding the large number of DoFs of lumped models and accurately captured the 
smooth bending deformation \cite{jones2006kinematics,Godage2011icra}. 
As these curve-parametric models only utilize the actual number of DoF (that is, length changes of PMAs), they are highly numerically efficient and physically accurate \cite{godage2015modal}.

The contribution of this paper is developing closed form forward and inverse kinematic models for an inextensible continuum arm based on the concept of constant curvature continuum sections built on PMAs. For the purpose, a set of curve parameters, detailed in \cite{godage2015modal}, are employed to map the muscle length changes in an over-constrained system to the spatial position and orientation of the continuum arm. The remaining of the paper is organized as follows. Section \ref{sec:prototype} explains the design and construction of the prototype arm as well as the materials utilized in the process. Section \ref{sec:FK} discusses the forward kinematic model of the arm. The closed form inverse kinematics of the arm is derived in Section \ref{sec:IK}. Section \ref{sec:simulation} is devoted to the verification of the inverse kinematics by a designed scenario. Section \ref{sec:concl} concludes the remarks and briefly introduces the future work.

\section{PROTOTYPE DESCRIPTION}\label{sec:prototype}
The proposed pneumatically-actuated soft arm prototype has two main components; the backbone and the pneumatic soft muscles. The backbone should allow the accommodation of the three pneumatic muscles, the fasten of both ends of the muscle, and the wrapping of the whole to maximize the bending capabilities and also stiffness range. The pneumatic soft muscles should have the extending capabilities in both directions, not be biased to any direction, and be able to sustain the pressure and return to its original form.

\subsection{Backbone Design} \label{subsec:backbone}
The backbone was inspired by the tails of some animals such as the spider monkey. As such, the continuum arm should be able to support itself (sufficient structural strength), and move fast, but with a soft, highly conformable environmental interface. The usefulness of this mode of operation becomes quite evident when we consider a highly adaptable biological example, such as the tail of spider monkeys; Fig. \ref{fig:fig1}(A). As a consequence of their structure, natural muscles have several mechanical properties which differ from conventional actuators. Muscles (and tendons) have spring-like properties, with inherent stiffness and damping. Thus, they are able to seamlessly transform between modes. For example, the tail of a monkey can act as a manipulator or supporting structure while standing upright, and a balancing appendage during jumping and climbing. This amazing capability is achieved while still being \textit{soft} to the touch because of the tail's muscular lining, which actuates the skeletal structure underneath; Fig. \ref{fig:fig1}(B). The fluid-based muscles of the proposed prototype, Fig. \ref{fig:fig1}(C), shoulder the actuation task while the backbone, Fig. \ref{fig:fig1}(D), offers structural integrity.

\subsection{Pneumatic Muscle Actuators}
The muscles that will provide the actuation for the arm are pneumatic and soft. The actuator is fabricated as follows \cite{godage2012pna}. The inner diameter and thickness of the silicone tube are, respectively, 9 and 2 mm respectively. The nylon mesh has a minimum and maximum diameter of 10 and 35 mm, respectively. The selected materials and a sample of the PMA are shown in Fig. \ref{fig:fig1}(D). The muscle is capable of contracting by $25\%$ and withstanding a range of pressures from 90 to 700 kPa.

\section{Curve Parametric Forward Kinematics}\label{sec:FK}
This section deals with deriving the curve parameters that define the spatial orientation of a single section contracting mode continuum arm. The forward kinematics of the arm is, subsequently, discussed.
\begin{figure}
  \centering
  \includegraphics[width=0.65\columnwidth]{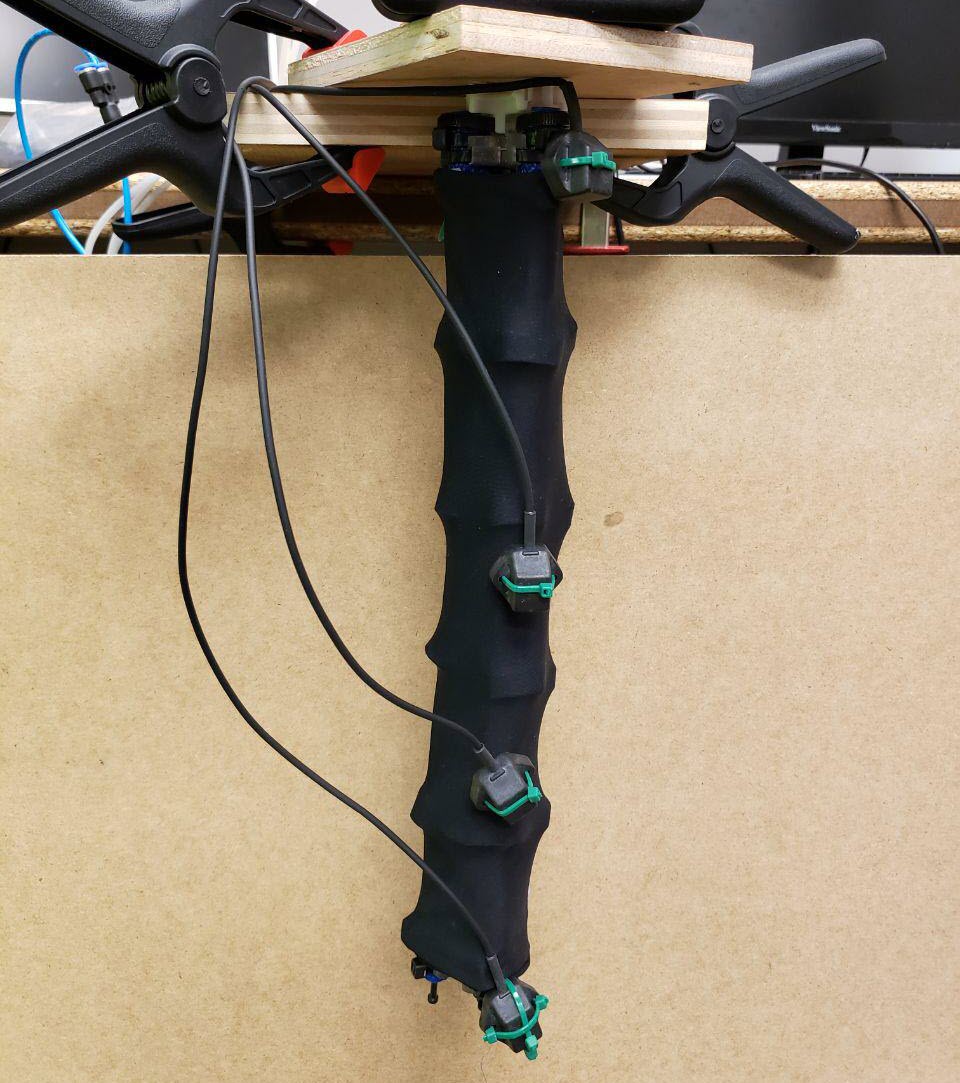}
  \caption{Single section inextensible continuum arm.}
  \label{fig:arm}
\end{figure}

\subsection{System Model}\label{subsec:SysModel}
\begin{figure}
\begin{centering}
\includegraphics[width=\columnwidth]{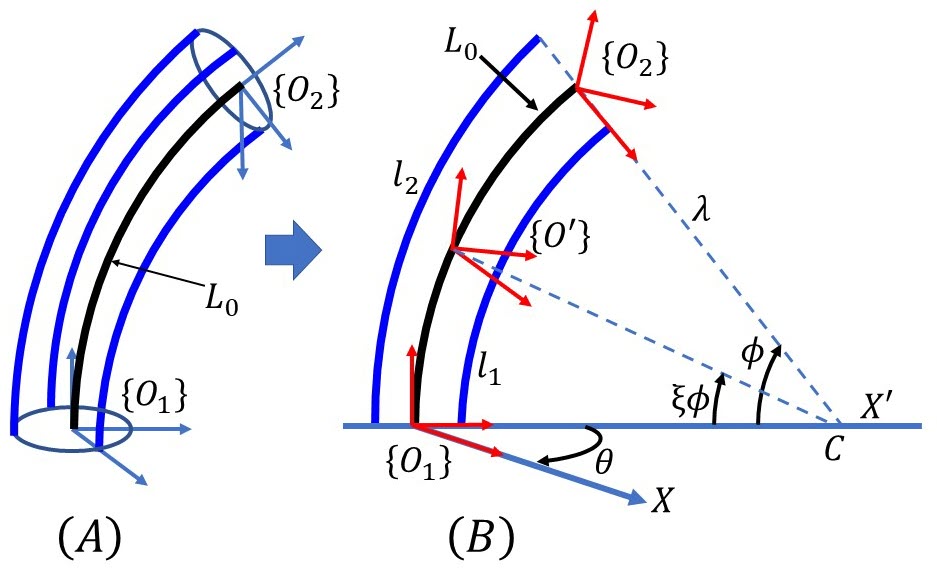}
\par\end{centering}
\caption{(A) Schematic of the continuum arm and its inextensible backbone, (B) Side view of the continuum segment.}
\label{fig:schematic}
\end{figure}

The single section continuum arm of the research, Fig. \ref{fig:arm}, consists of three mechanically identical contracting mode PMAs fixed to an inextensible rigid backbone. The schematic of the arm is shown in Fig. \ref{fig:schematic}.
The three PMAs are placed on a circular rigid frame at a radius $r$ from the center and $\frac{2\pi}{3}$ rads apart. The center point of the circular frame is thought of the center point of the backbone. Actuators are mechanically required to actuate at the distance of $r$ to the backbone. The actuation line is, then, parallel to the backbone axis; a line running through the backbone center along the length of the continuum arm. The actuators initial length is $L_{0}$, and the length variation is ${l_{i}} \in {\rm I\!R}$ where ${l_{i:\min }} \le {l_{i}}(t) \le {l_{i:\max }}$ for $i \in \{ 1,2,3\} $ and $t$ stand for the actuator number and time, respectively. Therefore, the length of each actuator at any time is calculated by ${L_{i}}(t) = L_{0} + {l_{i}}(t)$. The joint variable vector of the arm is also $\textbf{\textit{q}} = \left[ {{l_{1}}(t),\,{l_{2}}(t),\,{l_{3}}(t)} \right]^T$.

\subsection{Curve Parameters}\label{subsec:CPs}
Considering the backbone as a mechanical constraint, the continuum arm cannot exhibit any straight movement along with the axis of the backbone. However, it can bend in a circular arc shape given length change of actuators. Therefore, a circular arc with a variable curvature radius and length can represent the spatial orientation of the arm. The arc is described by three spatial parameters: radius of curvature, $\lambda \in (0,\infty)$, with instantaneous center $C$, angle subtended by the bending arc, $\phi \in [0,\pi)$, and angle of the bending plane with respect to the $+X$ axis (that is, angle between $\overrightarrow {{O_{1}}X}$ and $\overrightarrow {{O_{1}}C}$ in Fig. \ref{fig:schematic}), $\theta \in [-\pi,\pi]$. In order to derive the curve parameters, a geometrical approach, proposed by \cite{godage2015modal}, is adopted as follows.

\begin{figure}
\begin{centering}
\includegraphics[width=0.65\columnwidth]{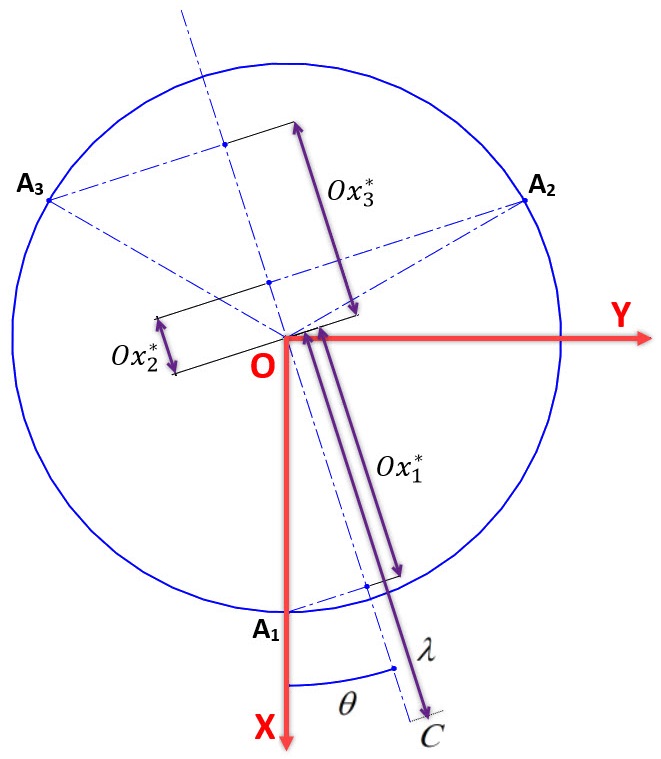}
\par\end{centering}
\caption{Projection of the actuation points of the arm.}
\label{fig:Ox_i}
\end{figure}

Suppose that the origin of the task space coordinate frame, $\{O\}$, coincides with the center of the base plane; therefore, $\overrightarrow {OX}$ marks $\overrightarrow {{OA}_{1}}$. Three actuation points $(A_{1}, A_{2}, A_{3})$ form an equilateral triangle of side $r\sqrt 3$. The coordinates of these points are $A_{1} = {[r,\,0,\,0]^T}$, $A_{2} = {\frac{r}{2}}{[-1,\,\sqrt 3,\,0]^T}$, and $A_{3} = {\frac{-r}{2}}{[1,\,\sqrt 3,\,0]^T}$. The circular arc representing the shape of the bent arm has an instantaneous center at $C$. Fig. \ref{fig:Ox_i} shows the projection of actuation points onto $\overrightarrow {OC}$, which intersect at ${x_{1}}^*$, ${x_{2}}^*$, and ${x_{3}}^*$. The corresponding distances between $O$ and intersection points are calculated by
\begin{align}
\begin{split}
{Ox_{1}}^{*} & = r\cos(\theta)\\
{Ox_{2}}^{*} & = r\cos\left(\frac{2\pi}{3} - \theta\right)\\
{Ox_{3}}^{*} & = r\cos\left(\frac{4\pi}{3} - \theta\right)
\end{split}
\label{eq:Ox*}
\end{align}
where $i \in \{ 1,2,3\}$. As shown in Fig. \ref{fig:schematic} and \ref{fig:Ox_i}, three actuator lengths correspond to the length of three concentric arcs. Considering the arc geometrical relationship, ${\rm{arc}}\,{\rm{length  =  curvature}}\,{\rm{radius}} \times {\rm{subtended}}\,{\rm{angle}}$, the actuator lengths are described in terms of curve parameters as
\begin{align}
\begin{split}
{L_{0}} + {l_{1}} & = (\lambda  - {Ox_{1}}^{*})\,\phi\\
 & = (\lambda  - r\cos(\theta))\,\phi
\end{split}
\label{eq:l1CP}
\end{align}
\begin{align}
\begin{split}
{L_{0}} + {l_{2}} & = (\lambda  - {Ox_{2}}^{*})\,\phi\\
 & = \left(\lambda  + {\frac{1}{2}} r\cos(\theta) - {\frac{\sqrt 3}{2}}r\sin(\theta)\right)\phi
\end{split}
\label{eq:l2CP}
\end{align}
\begin{align}
\begin{split}
{L_{0}} + {l_{3}} & = (\lambda  - {Ox_{3}}^{*})\,\phi\\
 & = \left(\lambda  + {\frac{1}{2}} r\cos(\theta) + {\frac{\sqrt 3}{2}}r\sin(\theta)\right)\phi
\end{split}
\label{eq:l3CP}
\end{align}

Since the continuum arm is constrained to the backbone, the arc length is constant; equal to the initial length of the identical muscles. Considering $\lambda \phi  = {L_{0}}$, length of the muscles, $l_{i}$ for $i \in \{ 1,2,3\}$, is derived by
\begin{align}
l_{i} & =-Ox_{i}^{*}\,\phi
\label{eq:l_i}
\end{align}
Plugging (\ref{eq:Ox*}) in (\ref{eq:l_i}) for $i \in \{ 1,2,3\}$ provides the length of actuators as
\begin{align}
{l_{1}} & = (-r\cos(\theta))\,\phi
\label{eq:l1}
\end{align}
\begin{align}
{l_{2}} & = \left({\frac{1}{2}} r\cos(\theta) - {\frac{\sqrt 3}{2}} r\sin(\theta)\right)\phi
\label{eq:l2}
\end{align}
\begin{align}
{l_{3}} & = \left({\frac{1}{2}} r\cos(\theta) + {\frac{\sqrt 3}{2}} r\sin(\theta)\right)\phi
\label{eq:l3}
\end{align}
Summing up (\ref{eq:l1}), (\ref{eq:l2}), and (\ref{eq:l3}) states that
\begin{align}
{l_{1}} + {l_{2}} + {l_{3}} = 0
\label{eq:lConstraint}
\end{align}
Therefore, there are just two independent joint variables, and the third one can be eliminated from the equations. To this end, ${l_{1}}$ in (\ref{eq:l1}), (\ref{eq:l2}), and (\ref{eq:l3}) is replaced by $(-l_{2} - l_{3})$.
Summing up (\ref{eq:l2}) and (\ref{eq:l3}) and rearranging the terms provides
\begin{align}
\cos(\theta) = {l_{2} + l_{3} \over r\phi}
\label{eq:costheta}
\end{align}
Subtracting (\ref{eq:l2}) from (\ref{eq:l3}) and rearranging the terms also gives
\begin{align}
\sin(\theta) = {l_{3} - l_{2} \over \sqrt 3 r\phi}
\label{eq:sintheta}
\end{align}
Applying (\ref{eq:costheta}) and (\ref{eq:sintheta}) to the trigonometric identity ${\sin ^2}\,(\theta)  + {\cos ^2}\,(\theta)  = 1$ and, then, solving for $\phi$ results in
\begin{align}
\phi\,(q) = \frac{2}{r} \sqrt {{l_{2}}^2 + {l_{3}}^2 + {l_{2}}{l_{3}} \over 3}
\label{eq:phi}
\end{align}
Dividing (\ref{eq:sintheta}) by (\ref{eq:costheta}) yields $\theta$ as
\begin{align}
\theta\,(q) = \tan^{-1} \left(l_{3} - l_{2} \over \sqrt 3\, (l_{2} + l_{3})\right)
\label{eq:theta}
\end{align}
Employing (\ref{eq:phi}) and considering $\lambda \phi  = {L_{0}}$, $\lambda$ is derived as
\begin{align}
\lambda\,(q) = \frac{r L_{0}}{2} \sqrt {3 \over {{l_{2}}^2 + {l_{3}}^2 + {l_{2}}{l_{3}}}}
\label{eq:lambda}
\end{align}
The spatial curve parameters $\lambda$, $\phi$, and $\theta$ are now expressed in terms of joint space variables.

\subsection{Forward Kinematics Model}\label{subsec:FK}
\begin{figure}
\begin{centering}
\includegraphics[width=\columnwidth]{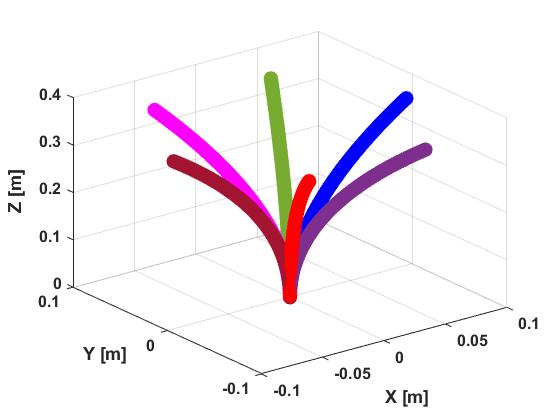}
\par\end{centering}
\caption{Illustration of the forward kinematics using different curve parameters.}
\label{fig:FK}
\end{figure}

Considering different orientations of the continuum section along its length, a moving coordinate frame, $\{O'\}$ in Fig. \ref{fig:schematic}, is defined. The moving frame can slide based on the value of a scalar parameter, $\xi \in [0,1]$, from the base ($\xi = 0$) to the tip ($\xi = 1$) of the continuum section along the backbone center line. Employing the curve parameters and considering the moving frame, the homogeneous transformation matrix (HTM), $\textbf{T} \in SE(3)$, is derived based on the approach elaborated in \cite{godage2015modal} by
\begin{align}
\begin{split}
\textbf{T}(c,\xi) & = {\textbf{R}_{z}}(\theta)\,{\textbf{P}_{x}}(\lambda)\,{\textbf{R}_{y}}(\xi \phi)\,{\textbf{P}_{x}}(-\lambda)\,{{\textbf{R}_{z}}^T}(-\theta) \\ 
&= \begin{bmatrix}
\textbf{R}(c,\xi) & \textbf{\textit{p}}(c,\xi) \\
\textbf{0}_{1\times3} & 1
\end{bmatrix}
\end{split}
\label{eq:HTM}
\end{align}
where $c = [\lambda,\,\phi,\,\theta]^T$ is the curve parameters vector, $\textbf{R}_{z} \in SO(2)$ and $\textbf{R}_{y} \in SO(2)$ are homogeneous rotation matrices about the $Z$ and $Y$ axes, $\textbf{P}_{x} \in {\rm I\!R}$ is the homogeneous translation matrix along the $X$ axis, and $\textbf{R} \in SO(3)$ and $\textbf{\textit{p}} = [x,\,y,\,z]^T$ are, respectively, the rotation and position matrices of the continuum arm in terms of curve parameters. The elements of $\textbf{R}$ and $\textbf{\textit{p}}$ are
\begin{align}
\begin{split}
\left[\,\textbf{R}\,\right]_{11} &= {\cos ^2}(\theta) \,\cos (\xi \phi) + {\sin ^2}(\theta) \\
\left[\,\textbf{R}\,\right]_{12} &= \cos(\theta) \,\sin(\theta) \, (\cos(\xi \phi) - 1) \\
\left[\,\textbf{R}\,\right]_{13} &= \cos(\theta) \,\sin(\xi \phi) \\
\left[\,\textbf{R}\,\right]_{21} &= \left[\,\textbf{R}\,\right]_{12} \\
\left[\,\textbf{R}\,\right]_{22} &= {\sin ^2}(\theta) \,\cos(\xi \phi) + {\cos ^2}(\theta) \\
\left[\,\textbf{R}\,\right]_{23} &= \sin(\theta) \,\sin(\xi \phi) \\
\left[\,\textbf{R}\,\right]_{31} &= -\left[\,\textbf{R}\,\right]_{13} \\
\left[\,\textbf{R}\,\right]_{32} &= -\left[\,\textbf{R}\,\right]_{23} \\
\left[\,\textbf{R}\,\right]_{33} &= \cos(\xi \phi) \\
\left[\,\textbf{\textit{p}}\,\right]_1 &= \lambda \,\cos(\theta) \,(1 - \cos(\xi \phi)) \\
\left[\,\textbf{\textit{p}}\,\right]_2 &= \lambda \,\sin(\theta) \,(1 - \cos(\xi \phi)) \\
\left[\,\textbf{\textit{p}}\,\right]_{3} &= \lambda \,\sin(\xi \phi)
\end{split}
\label{eq:HTMelements}
\end{align}
Finally, the spatial coordinate of the arm tip and also every point along the backbone axis is calculated by the derived elements. Fig. \ref{fig:FK} shows the continuum arm bending toward different directions in the task space.

\begin{figure}
\begin{centering}
\includegraphics[width=\columnwidth]{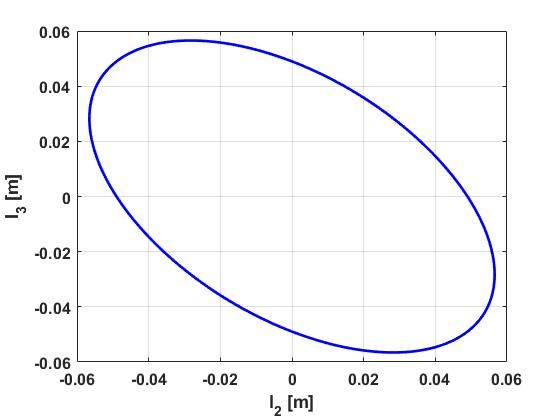}
\par\end{centering}
\caption{Valid combinations of $l_{2}$ and $l_{3}$.}
\label{fig:ellipse}
\end{figure}

As stated in \cite{godage2015modal}, the length variation of the actuators in a backboneless continuum arm should be ${l_{i:\min }} \le {l_i}(t) \le {l_{i:\max }}$ for the actuator number of $i \in \{ 1,2,3\}$. In contrast, in our inextensible continuum arm, considering the range of $\theta$ and $\phi$, the minimum and maximum values of the length of the actuators are calculated using (\ref{eq:l2}) and (\ref{eq:l3}). Since $l_{1}$ is the dependent joint variable, valid combinations of independent joint variables, $l_{2}$ and $l_{3}$, are shown in Fig. \ref{fig:ellipse}. The possible combinations are those inside the figured ellipse.

\section{Curve Parametric Inverse Kinematics}\label{sec:IK}
This section aims at deriving the closed-form inverse kinematics of the arm using the position of the arm tip. Suppose that the position of the arm tip is given as $(P_{x},P_{y},P_{z})$. According to (\ref{eq:HTMelements}), the arm tip position depends on the curve parameters as
\begin{subequations}
\begin{align}
P_{x} &= \frac{L_{0}}{\phi} \left\{\cos(\theta) \,(1 - \cos(\xi \phi)) \right\}\label{eq:Px}\\
P_{y} &= \frac{L_{0}}{\phi} \left\{\sin(\theta) \,(1 - \cos(\xi \phi)) \right\}\label{eq:Py}\\
P_{z} &= \frac{L_{0}}{\phi} \,\sin(\xi \phi)\label{eq:Pz}
\end{align}
\end{subequations}
Dividing (\ref{eq:Py}) by (\ref{eq:Px}) provides $\theta$ in terms of the arm tip position as
\begin{align}
\theta = tan^{-1} \left(\frac{P_y}{P_x}\right)
\label{eq:theta_IK}
\end{align}
Plugging (\ref{eq:theta_IK}) into (\ref{eq:Py}) states that
\begin{align}
\cos(\xi \phi) &= 1 - {P_{y}\phi \over F_{1}}
\label{eq:P_y2}
\end{align}
where $F_{1} = L_{0}\,\sin\left(tan^{-1}(\frac{P_{y}}{P_{x}})\right)$. Applying (\ref{eq:P_y2}) and rearranged form of (\ref{eq:Pz}) to the trigonometric identity ${\sin ^2}\,(\xi \phi)  + {\cos ^2}\,(\xi \phi)  = 1$ and, then, solving for $\phi$ results in
\begin{align}
\phi &= {2P_{y} \over F_{1} F_{2}}
\label{eq:phi_IK}
\end{align}
where $F_{2} = \left( \frac{P_{z}^2}{L_{0}^2} + \frac{P_{y}^2}{F_{1}^2}\right)$. Therefore, curve parameters can be calculated using the arm tip position.

\section{Simulation Results}\label{sec:simulation}
In this section, the inverse kinematic model developed in the previous section is employed to simulate how curve parameters and muscle lengths change during an intended kinematic maneuver.
The mechanical properties considered for the purpose are $L_{0}=0.37\,(m)$, $r=0.018\,(m)$, and $\xi=1$.

\begin{figure}
\begin{centering}
\includegraphics[width=\columnwidth]{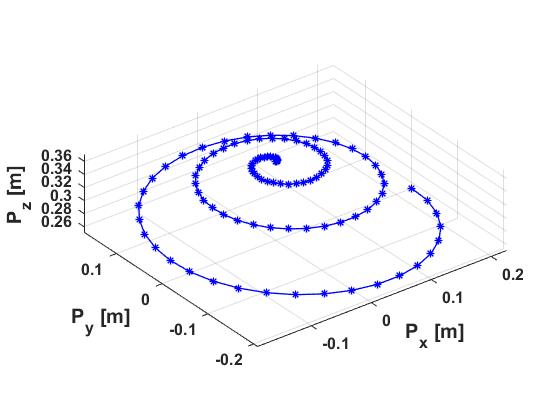}
\par\end{centering}
\caption{Spiral path generated for the inverse kinematics maneuver.}
\label{fig:spiral}
\end{figure}

First, a path is generated in the task space of the arm. The path is a spiral projected on a spherical shell which starts from the top highest point of the shell. Fig. \ref{fig:spiral} shows the intended path. The three axes of $P_x$, $P_y$, and $P_z$ in the figure are the position of the arm tip with respect to $x$, $y$, and $z$ axis, respectively. The significance of the path is that since it is three-dimensional, it requires all PMAs to be actuated during the path following. Also, it is highly similar to the paths that the arm should follow in typical tasks for which we will use the arm in future works. Then, the spatial position of the arm tip is fed into the kinematic model to find the profile of the curve parameters $\theta_{IK}$ and $\phi_{IK}$ as well as the muscle lengths $l_{2}$ and $l_{3}$ while the arm tip is following the given path.

\begin{figure}
\begin{centering}
\includegraphics[width=\columnwidth]{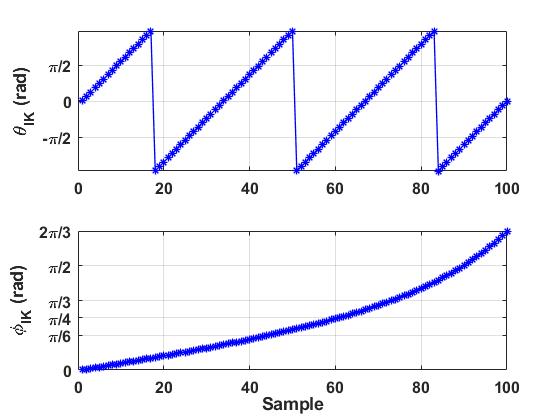}
\par\end{centering}
\caption{Profile of the curve parameters during the kinematic maneuver.}
\label{fig:thetaIKphiIK}
\end{figure}

\begin{figure}
\begin{centering}
\includegraphics[width=\columnwidth]{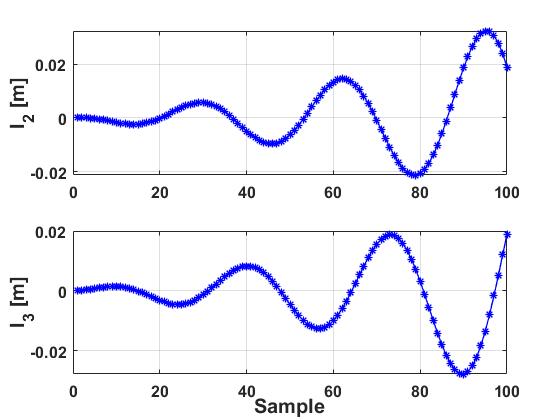}
\par\end{centering}
\caption{Profile of the muscle lengths during the kinematic maneuver.}
\label{fig:l2IKl3IK}
\end{figure}

\begin{figure}
\begin{centering}
\includegraphics[width=\columnwidth]{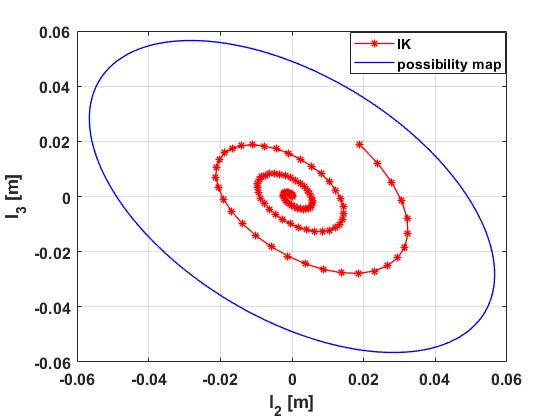}
\par\end{centering}
\caption{A comparison on the muscle lengths in the inverse kinematics and the possibility map.}
\label{fig:liMix}
\end{figure}

The profile of the curve parameters is shown in Fig. \ref{fig:thetaIKphiIK}. Fig. \ref{fig:l2IKl3IK} also shows the profile of the muscle lengths during the task. In order to prove that the final results of the kinematic maneuver are reliable, the muscle lengths obtained from the inverse kinematics are compared to the possibility map of the corresponding muscle lengths presented in Fig. \ref{fig:ellipse}. The comparative demonstration of the inverse kinematics and the possibility map is shown in Fig. \ref{fig:liMix}.

\section{CONCLUSION AND FUTURE WORKS}\label{sec:concl}
Inspired by the highly dexterous tails of Capuchin monkeys, a novel continuum arm design was introduced which uses both soft and rigid elements. We utilized a highly redundant rigid-discrete link chain as the backbone and PMAs for powering the continuum arm design. The use of backbone with the antagonistic arrangement of muscle actuators facilitates higher payloads (suitable for object manipulation) and decoupled stiffness and position control. Then, a curve parametric kinematic model was introduced for a contracting-mode inextensible continuum section. The closed form inverse kinematics of the arm was also derived. In order to verify the closed-form model, a spiral path in the task space was generated and fed to the inverse kinematic model. The result was, then, compared to the possibility map of the joint space variables, which showed the reliability of the developed model.

The future work will focus on the improvement of the kinematic model to consider uncertainties of the real world and also the torsion of the arm during tasks. Also, the closed form inverse kinematics of a multi-section inextensible arm will be the second potential research problem to construct continuum arms for human-friendly applications. We will also carry out an in-depth study on the stiffness control capabilities and decoupled stiffness and position control of the arm.


\bibliographystyle{IEEEtran}
\bibliography{refs}

\end{document}